\documentclass{SCIS2021}
\usepackage{listings}

\usepackage{booktabs}
\usepackage{graphics}
\usepackage{graphicx}
\usepackage{float}

\usepackage[T1]{fontenc}
\usepackage[scaled]{beramono}

\newcommand{\ie}{\emph{i.e.}}
\newcommand{\name}{{\bf PyCIL }}
\newcommand{\mame}{{\bf PyCIL}}
\newcommand{\x}{{\bf x}}

\newcommand{\D}{\mathcal{D}}

\newcommand{\R}{\mathbb{R}}

\begin{document}
%\oa
%%%%%%%%%%%%%%%%%%%%%%%%%%%%%%%%%%%%%%%%%%%%%%%%%%%%%%%
%%% Authors do not modify the information below
%%% 作者不需要修改此处信息
\ArticleType{LETTER}
%\SpecialTopic{}
\Year{2020}
\Month{}
\Vol{}
\No{}
\DOI{}
\ArtNo{}
\ReceiveDate{}
\ReviseDate{}
\AcceptDate{}
\OnlineDate{}
%%%%%%%%%%%%%%%%%%%%%%%%%%%%%%%%%%%%%%%%%%%%%%%%%%%%%%%

%%% title: 标题
%%%   \title{title}{title for citation}
\title{PyCIL: A Python Toolbox for \\Class-Incremental Learning}{PyCIL: A Python Toolbox for Class-Incremental Learning}

%%% Corresponding author: 通信作者
%%%   \author[number]{Full name}{{email@xxx.com}}
%%% General author: 一般作者
%%%   \author[number]{Full name}{}
\author{Da-Wei Zhou}{}
\author{Fu-Yun Wang}{}
\author{Han-Jia Ye}{{yehj@lamda.nju.edu.cn}}
\author{De-Chuan Zhan}{}

%%% Author information for page head. 页眉中的作者信息
\AuthorMark{Author A}

%%% Authors for citation. 首页引用中的作者信息
\AuthorCitation{Zhou D-W, Wang F-Y, Ye H-J, et al}

%%% Authors' contribution. 同等贡献
%\contributions{Authors A and B have the same contribution to this work.}

%%% Address. 地址
%%%   \address[number]{Affiliation, City {\rm Postcode}, Country}
\address{State Key Laboratory for Novel Software Technology
	Nanjing University, Nanjing 210023, China}
%\address[2]{Affiliation, City {\rm 000000}, Country}
%\address[3]{Affiliation, City {\rm 000000}, Country}

\maketitle

%%%%%%%%%%%%%%%%%%%%%%%%%%%%%%%%%%%%%%%%%%%%%%%%%%%%%%%
%%% The main text. 正文部分
%%%%%%%%%%%%%%%%%%%%%%%%%%%%%%%%%%%%%%%%%%%%%%%%%%%%%%%
\begin{multicols}{2}
%\deareditor
%Traditional machine learning systems are deployed under the closed-world setting, which requires the entire training data before the offline training process. However, real-world applications often face the incoming new classes, and a model should incorporate them continually. The learning paradigm is called Class-Incremental Learning (CIL). We propose a Python toolbox that implements several key algorithms for class-incremental learning to ease the burden of researchers in the machine learning community. The toolbox contains implementations of a number of founding works of CIL such as EWC~\cite{kirkpatrick2017overcoming} and iCaRL~\cite{rebuffi2017icarl}, but also provides current state-of-the-art algorithms that can be used for conducting novel fundamental research.
%This toolbox, named \name for Python Class-Incremental Learning, is available at \url{https://github.com/G-U-N/PyCIL}.
%\lettersection{Introduction and Motivations}
With the rapid development of deep learning, current deep models can learn a fixed number of classes with high performance. However, in our ever-changing world, data often comes from the open environment, which is  with stream format or available temporarily due to privacy issues.
As a result, the classification model should learn new classes incrementally instead of restarting the training process. A straightforward approach is to finetune the model with the incoming new data, while it suffers \emph{catastrophic forgetting} phenomena: due to the absence of previous data, the prediction on former classes drastically drops. Class-incremental learning (CIL) aims to extend the acquired knowledge with only new classes. 
For example, when training a robot in the open-world, it meets new objects as time goes by,
and in the electronic commerce platform, new types of products appear daily. 
%Figure~\ref{figure:incremental} demonstrates the setting of CIL. 
We give an example to demonstrate the setting of CIL.
In the first task, the model needs to classify birds and dogs. After that, the model is incrementally updated with two new classes, \ie, tigers and fish, and it needs to classify among two old classes (birds and dogs) and two new classes (tigers and fish). Similarly, new classes like monkeys and sheep will emerge in the next task, requiring the model to incorporate them incrementally.
New categories arrive progressively, and the model needs to classify more classes without forgetting the former ones.

%Figure~\ref{figure:incremental} demonstrates the setting of CIL. 
%In the first task, the model needs to classify birds and dogs. After that, the model is incrementally updated with two new classes, \ie, tigers and fish, and it needs to classify among two old classes (birds and dogs) and two new classes (tigers and fish). Similarly, new classes like monkeys and sheep will emerge in the next task, requiring the model to incorporate them incrementally.
%New categories arrive progressively, and the model needs to classify more classes without forgetting the former ones.

With the growing interest of the machine learning community in class-incremental learning, it is essential to provide a simple and efficient toolbox with several class-incremental learning algorithms. We choose to conduct its development in the Python programming language for its wide use in the machine learning community. Its high-level interactive nature makes it an appealing tool for both academic and industrial software developments, and several popular machine learning libraries and deep learning open source frameworks are built upon it.

The Python Class-Incremental Learning (\mame) library takes advantage of Python to make Class-Incremental Learning accessible to the machine learning community. 
It contains implementations of several founding works of CIL and provides current state-of-the-art algorithms that can be used to conduct novel fundamental research.
As \name is designed to be user-focused and friendly, we have kept our toolbox easy to use and accessible with convention consistencies and syntax over all the available functions. Moreover, our toolbox depends only on standard open-source libraries, and it is usable under many operating systems such as Linux, MacOSX, or Windows. The source code of \name is available at \url{https://github.com/G-U-N/PyCIL}.

%\begin{figure}[H]
%	\centering
%	\includegraphics[width=1\columnwidth]{./pics/teaser_eng}
%	\vspace{-5mm}
%	\caption{  The setting of CIL. Non-overlapping classes arrive sequentially, and we need to build a classifier for all the classes incrementally. We only have access to the classes of the current task and need to learn the new classes with the current model. After the learning stage of each task, the model is evaluated among all seen classes, \ie, it should not only perform well in the newly learned classes but also remember the former  classes without catastrophic forgetting.
%	} \label{figure:incremental}
%	\vspace{-4mm}
%\end{figure}

\definition[Class-Incremental Learning] Class-incremental learning was proposed to learn a stream of data incrementally from different classes. Assume there are a sequence of $B$  training tasks $\left\{\D^{1}, \D^{2}, \cdots, \D^{B}\right\}$ without overlapping classes, where $\D^{b}=\left\{\left(\x_{i}^{b}, y_{i}^{b}\right)\right\}_{i=1}^{n_b}$ is the $b$-th incremental step with $n_b$ instances. Besides,  $\x_i^b \in \R^D$ is a training instance of class $y_i \in Y_b$, $Y_b$ is the label space of task $b$, where
%Generally, there are no class overlapping between tasks, \ie, 
$Y_b  \cap Y_{b^\prime} = \varnothing$ for $b\neq b^\prime$. 
During the training process of task $b$, we can only access data from $\D^b$.  
The aim of CIL at each step is not only to acquire the knowledge from the current task  $\D^b$, but also to preserve the knowledge from former tasks. 
After each task, the trained model is evaluated over all seen classes $\mathcal{Y}_b=Y_1 \cup \cdots Y_b$.

\begin{figure*}[t]
	
	\centering
	\subfloat[CIFAR100, 10 Stages]{
		\includegraphics[width=0.49\columnwidth]{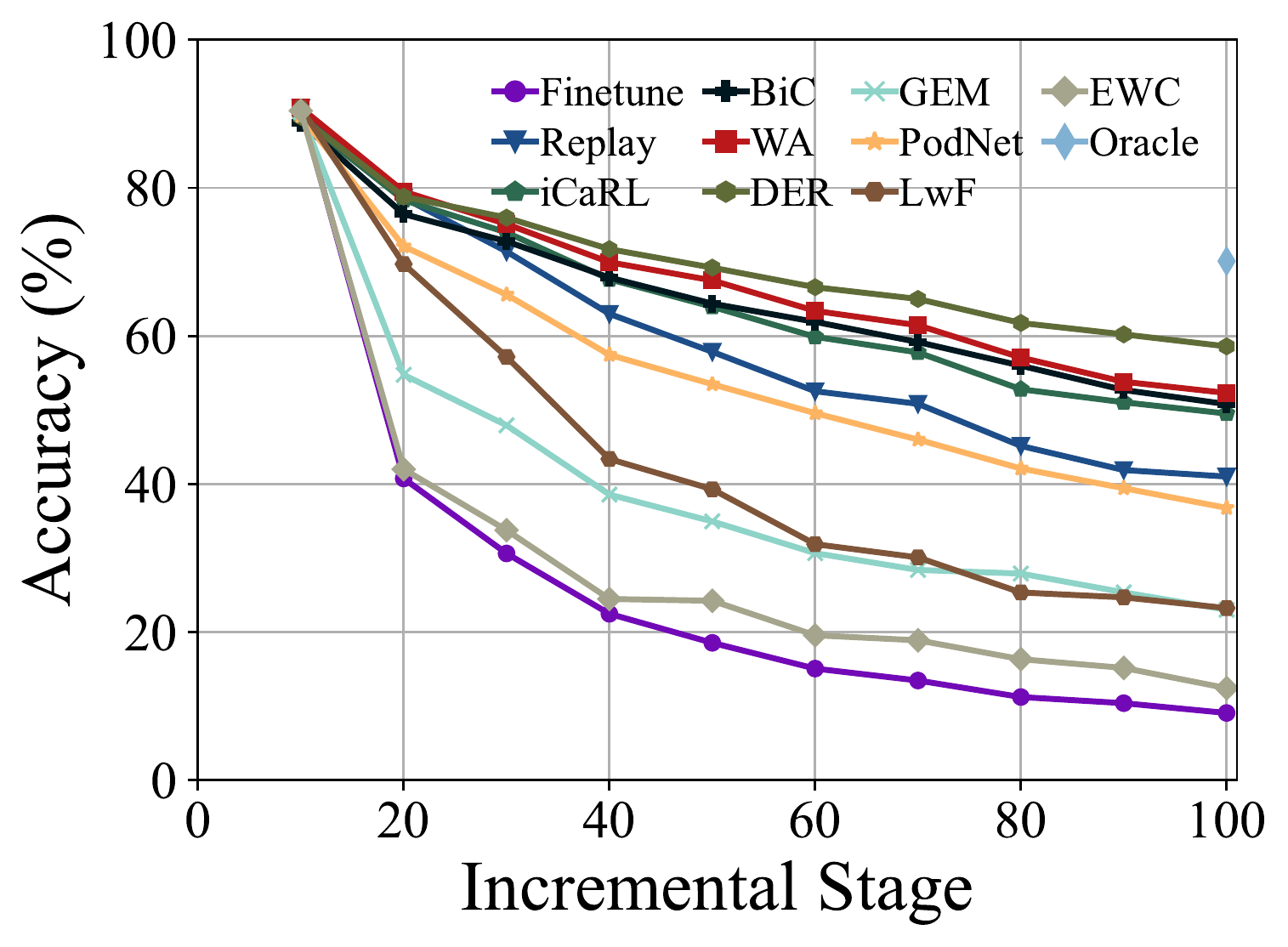}} 
	\subfloat[CIFAR100, B50, 5 Stages]{
		\includegraphics[width=0.49\columnwidth]{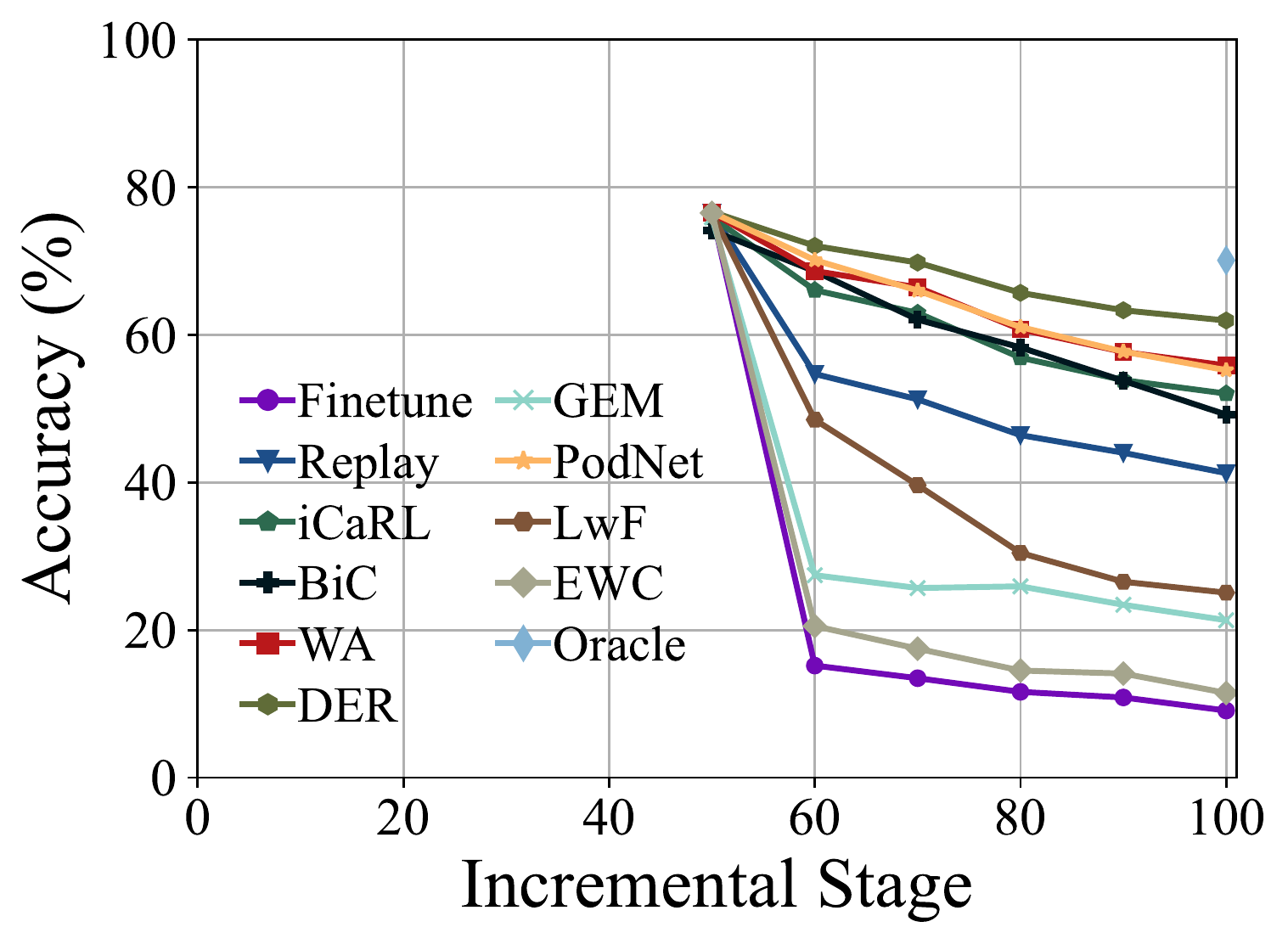}} 
	\subfloat[ImageNet100, 10 Stages]{
		\includegraphics[width=0.49\columnwidth]{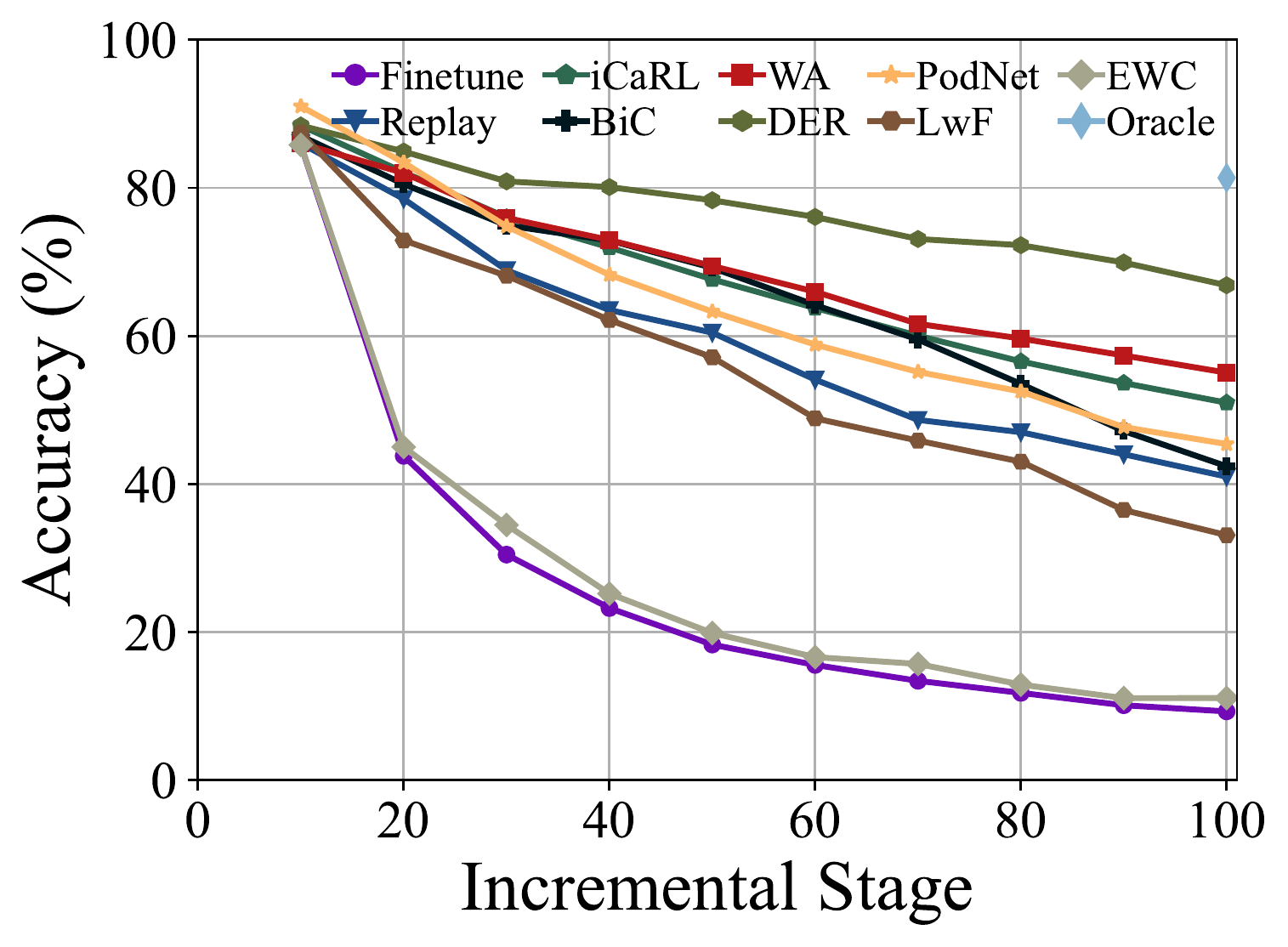}} 
	\subfloat[ImageNet100, B50, 5 Stages]{
		\includegraphics[width=0.49\columnwidth]{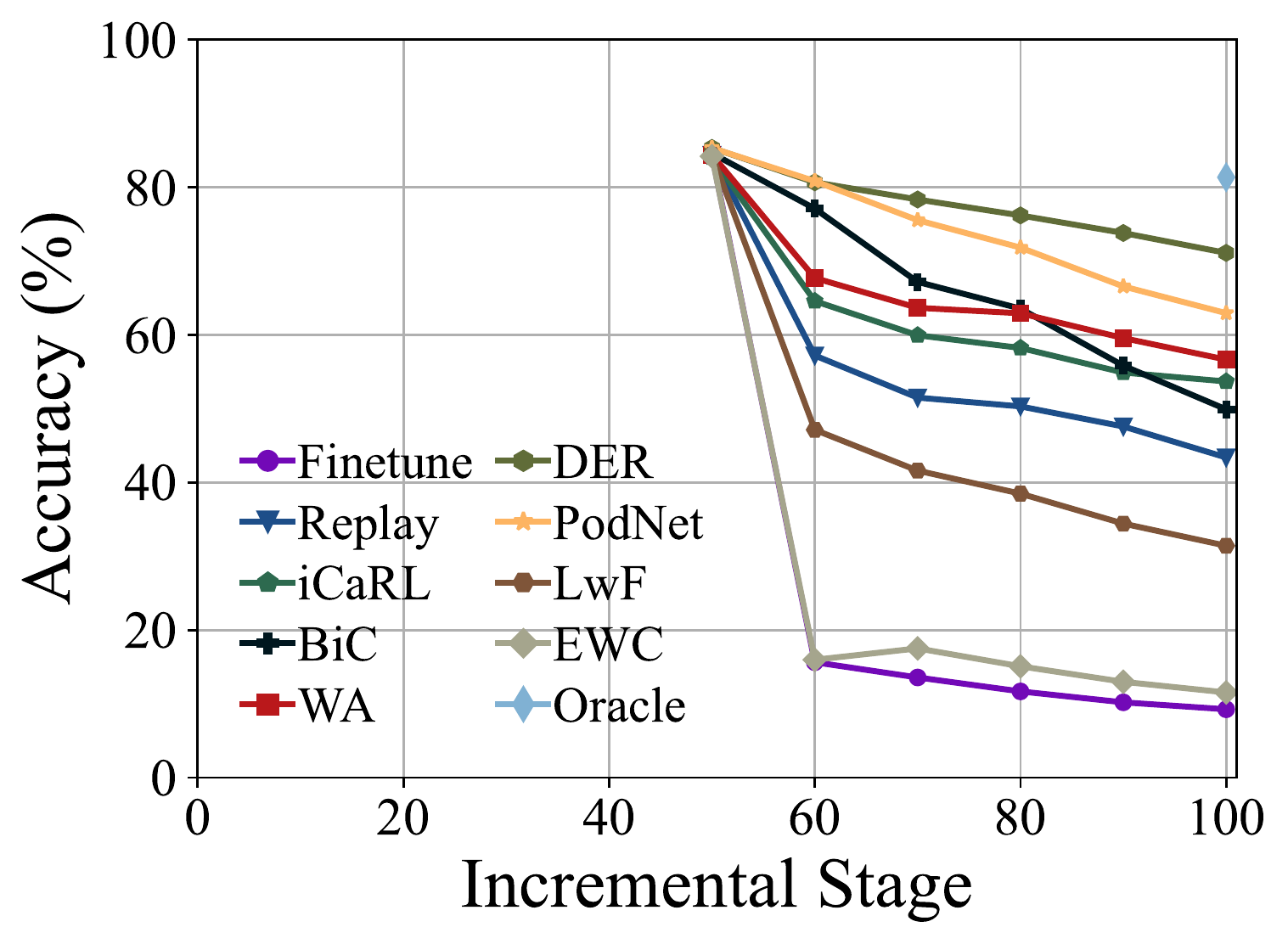}} 
%	\vspace{-3mm}
	\caption{  Reproduced incremental accuracy on CIFAR100 and ImageNet100. 
	} 
%	\vspace{-3mm}
	\label{figure:results}
	
\end{figure*}

\definition[Exemplar Set] In the $b$-th stage, typical CIL methods update the model with only the current dataset $\D^{b}$, which suffers severe catastrophic forgetting. As a result, current CIL methods propose to maintain an extra exemplar set $\mathcal{E}=\left\{\left(\x_{j}, y_{j}\right)\right\}_{j=1}^{M}$. $\mathcal{E}$ helps to reserve a limited amount of instances for the classes seen before, and revisiting them can help the model overcome catastrophic forgetting. The exemplars are selected with the herding algorithm to make them more representative.

%Class-incremental learning was proposed to learn a stream of data incrementally from different classes. Assume there are a sequence of $B$  training tasks $\left\{\D^{1}, \D^{2}, \cdots, \D^{B}\right\}$ without overlapping classes, where $\D^{b}=\left\{\left(\x_{i}^{b}, y_{i}^{b}\right)\right\}_{i=1}^{n_b}$ is the $b$-th incremental step with $n_b$ instances. Besides,  $\x_i^b \in \R^D$ is a training instance of class $y_i \in Y_b$, $Y_b$ is the label space of task $b$, where
%%Generally, there are no class overlapping between tasks, \ie, 
%$Y_b  \cap Y_{b^\prime} = \varnothing$ for $b\neq b^\prime$. 
%During the training process of task $b$, we can only access data from $\D^b$.  
%The aim of CIL at each step is not only to acquire the knowledge from the current task  $\D^b$, but also to preserve the knowledge from former tasks. 
%After each task, the trained model is evaluated over all seen classes $\mathcal{Y}_b=Y_1 \cup \cdots Y_b$.

\lettersection{Implemented Algorithms}
In \name, we implemented 11 typical algorithms for class-incremental learning. They are listed as:
	 {\bf Finetune:} The baseline method which simply updates parameters on new tasks and suffers from
	severe catastrophic forgetting. 
	 {\bf Replay:} The baseline method which updates parameters on new tasks with instances from the new dataset and exemplar set.
	 {\bf EWC~\cite{kirkpatrick2017overcoming}:} Uses Fisher Information Matrix to weigh the importance of each parameter and regularizes them to overcome forgetting.
	 {\bf LwF~\cite{li2017learning}:} Uses knowledge distillation to align the output probability between old and new models.
	 {\bf iCaRL~\cite{rebuffi2017icarl}:} Based on LwF, it  introduces an exemplar set for rehearsal and uses the nearest center mean classifier.
	 {\bf GEM~\cite{lopez2017gradient}:} Uses exemplars as the regularization of gradient updating.
	 {\bf BiC~\cite{wu2019large}:} Trains an extra adaptation layer based on iCaRL, which adjusts the logits on new classes.
	 {\bf WA~\cite{zhao2020maintaining}:} Normalizes the classifier weight after each learning session based on iCaRL. 
	 {\bf PODNet~\cite{douillard2020podnet}:} Introduces pooled outputs distillation to constrain the network.
	 {\bf DER~\cite{yan2021dynamically}:} A two-stage learning approach that utilizes a dynamically expandable representation for more effective incremental 	concept modeling.
	 {\bf Coil~\cite{zhou2021co}:} Builds bi-directional knowledge transfer in the incremental learning process with optimal transport. 
	 %It first addresses the ability that old model can help learning new classes.

{\noindent\bf Dependencies:} \name relies on open source libraries such as NumPy and SciPy for linear algebra and optimization problems. The network structure is designed with  PyTorch.

%{\noindent\bf Basic Usage:} \name provides implementations of the above 11 methods. As for the benchmark dataset setting in class-incremental learning, we provide the  environment of CIFAR100 and ImageNet100/1000. 
%When using \mame, users can edit the global parameters and algorithm-specific hyper-parameter, and then run: 
%python main.py --config=./exps/[MODEL NAME].json
%where MODEL NAME should be chosen from the above 11 methods. The aforementioned global parameters include: {\bf Memory-Size:} The total exemplar number in the incremental learning process. Assuming there are $K$ classes at current stage, the model will preserve $\left[\frac{{memory-size}}{K}\right]$ exemplar per class.	 {\bf Init-Cls:} The number of classes in the first incremental stage. Since there are different settings in CIL with a different number of classes in the first stage, our framework enables different choices to define the initial stage. {\bf Increment:} The number of classes in each incremental stage $b, b>1$. By default, the number of classes per incremental stage is equivalent per stage. {\bf Convnet-type:} The backbone network for the incremental model. In the benchmark-setting, ResNet32 is utilized for CIFAR100, and ResNet18 is utilized for ImageNet. {\bf Seed:} The random seed for shuffling the class order, which is set to 1993 by default.

{\noindent\bf Basic Usage:} \name provides implementations of the above 11 methods. As for the benchmark dataset setting in class-incremental learning, we provide the  environment of CIFAR100 and ImageNet100/1000. 
When using \mame, users can edit the global parameters and algorithm-specific hyper-parameter, and then run the main function. The aforementioned global parameters include: {\bf Memory-Size:} The total exemplar number in the incremental learning process. {\bf Init-Cls:} The number of classes in the first incremental stage. {\bf Increment:} The number of classes in each incremental stage $b, b>1$. {\bf Convnet-type:} The backbone network for the incremental model. {\bf Seed:} The random seed for shuffling the class order, which is set to 1993 by default.

\lettersection{Evaluation} The common performance measure for CIL is the test accuracy after every stage, denoted as $\mathcal{A}_b$, where $b$ is the stage index. Similarly, the averaged accuracy across all stages is also a common measure, \ie, $\bar{\mathcal{A}}=\frac{1}{B} \sum_{b=1}^{B} \mathcal{A}_{b}$. As a preliminary step for research in the machine learning field, we have tested the incremental performance (Top-1 accuracy) along the incremental stages, and the results are shown in Figure~\ref{figure:results}. We use the benchmark datasets, \ie, CIFAR100 and ImageNet100, and divide the 100 classes into several incremental stages. 
%We also report the averaged incremental accuracy for all stages in Table~\ref{tab:avg-acc} for CIFAR100 dataset. 
Since some parameters are not reported in the original paper, we search for a good parameter set in our re-implementation.
Most reproduced algorithms have the same or even better performance than the results reported in the original paper.

%\begin{table}[H]
%	\caption{Comparison of reproduced and reported average accuracy on 10-stage CIFAR100.}
%	\label{tab:avg-acc}
%	\footnotesize
%	%\tabcolsep 20pt 
%	\centering
%	\begin{tabular*}{0.98\columnwidth}{cccc}
%		%		\toprule
%		\hline
%		Method & Reproduced &  Reported&Running Time (Min)\\
%		\hline		%\midrule
%		%Finetune & 26.25 & - \\
%		%Replay & 59.31 & -  \\
%		%GEM & 40.18 & - \\
%		%LwF & 43.56 & - \\
%		iCaRL & 64.42 & 64.10 & 182.1\\
%		%EWC & 29.73 &  - \\
%		WA & 67.09 & 		64.5 & 184.5\\
%		%PODNet & 55.22 & - \\
%		%BiC & 65.08 & - \\
%		Coil & 65.48 & 65.48 & 409.4\\
%		DER &  69.74 &   69.41 & 537.0 \\
%		%		\bottomrule 
%		\hline
%		\vspace{-8mm}
%	\end{tabular*}
%\end{table}

\lettersection{Conclusion}
We have presented \mame, a class-incremental learning toolbox written in Python. It contains implementations of a number of founding works of CIL, but also provides current state-of-the-art algorithms that can be used to conduct novel fundamental research.
Code consistency makes it an easy tool for research purposes, teaching, and industrial applications.

\Acknowledgements{This research was supported by National Key R\&D
	Program of China (2020AAA0109401), NSFC (61773198,	61921006, 62006112), NSFC-NRF Joint Research Project
	under Grant 61861146001, Collaborative
	Innovation Center of Novel Software Technology and Industrialization, NSF of Jiangsu Province (BK20200313). Da-Wei Zhou and Fu-Yun Wang have the equal contributions.}

%%%%%%%%%%%%%%%%%%%%%%%%%%%%%%%%%%%%%%%%%%%%%%%%%%%%%%%
%%% Supplements. 补充材料, 非必选
%%%%%%%%%%%%%%%%%%%%%%%%%%%%%%%%%%%%%%%%%%%%%%%%%%%%%%%
%\Supplements{Appendix A.}

%%%%%%%%%%%%%%%%%%%%%%%%%%%%%%%%%%%%%%%%%%%%%%%%%%%%%%%
%%% Reference section. 参考文献
%%% citation in the content using "some words~\cite{1,2}".
%%% ~ is needed to make the reference number is on the same line with the word before it.
%%% Please make sure there are no more than 9 items of references.
%%%%%%%%%%%%%%%%%%%%%%%%%%%%%%%%%%%%%%%%%%%%%%%%%%%%%%%
%\begin{thebibliography}{99}
%
%\bibitem{1} Author A, Author B, Author C. Reference title. Journal, Year, Vol: Number or pages
%
%\bibitem{2} Author A, Author B, Author C, et al. Reference title. In: Proceedings of Conference, Place, Year. Number or pages
%
%\end{thebibliography}
\bibliography{scis}
\bibliographystyle{unsrt}
\end{multicols}
\end{document}